\documentclass[preprint,12pt]{elsarticle}

\usepackage{algorithm}
\usepackage{algorithmic}
\usepackage{subcaption}
\usepackage{caption}
\usepackage{adjustbox}
\usepackage{longtable}
\usepackage{makecell}
\usepackage{float}
\usepackage{rotating}
\usepackage{booktabs}
\usepackage{bm}
\usepackage{xcolor}

\usepackage{amssymb}
\usepackage{amsmath}
\usepackage{graphicx}
\graphicspath{{figures/}}

\journal{Nuclear Physics B}

\begin{document}

\begin{frontmatter}



\title{Meta-Black-Box Optimization with Ensemble Surrogate Modeling for Robustness–Accuracy Trade-off within SAEA}

\author[label1]{Xiao Jin}
\author[label1]{Yongxiong Wang*}
\author[label2]{Haobo Liu}
\author[label3]{Yudong Du}
\author[label2]{Yukun Du*} 

\affiliation[label1]{organization={School of Optical-Electrical and Computer Engineering, University of Shanghai for Science and Technology},
            postcode={200093}, 
            city={Shanghai},
            country={China}}
\affiliation[label2]{organization={College of System Engineering, National University of Defense Technology},
            postcode={410073}, 
            city={Changsha},
            country={China}}

\affiliation[label3]{organization={Xinxiang University},
            postcode={453003}, 
            city={Xinxiang},
            country={China}}

\begin{abstract}
Surrogate-Assisted Evolutionary Algorithms (SAEAs) have been widely used for expensive black-box optimization problems. However, their reliance on rigid and manually designed components limits flexibility and generalization across tasks. Meta-Black-Box Optimization (MetaBBO) provides a promising solution by adaptively configuring algorithmic components. Yet, existing MetaBBO approaches typically control only a single component, and few studies address the unified control of multi-component optimizers like SAEA. Moreover, current research rarely considers the robustness–accuracy trade-off in surrogate modeling, which is essential for stable early exploration and accurate late-stage exploitation. To address these issues, we propose AdaE-SAEA (Adaptive Ensemble Surrogate-Assisted Evolutionary Algorithm) for expensive multi-objective optimization. AdaE-SAEA integrates SAEA as the low-level optimizer within the MetaBBO framework and jointly controls both the infill criterion and ensemble-based surrogate modeling. Bagging and boosting are designed as surrogate modeling modules to adaptively balance robustness and accuracy during different search phases, while the meta-policy simultaneously controls the infill criterion to ensure adaptive sampling decisions. The meta-policy is trained via reinforcement learning with parallel sampling and centralized training for improved efficiency and transferability. Experiments on synthetic and real-world problems show that AdaE-SAEA outperforms state-of-the-art baselines and MetaBBO-based methods. Additionally, we demonstrate the effectiveness of TabPFN as the base surrogate model for ensemble learning. To the best of our knowledge, this is the first work to unify the control of surrogate modeling and infill criteria in SAEA while explicitly addressing the robustness–accuracy trade-off.
\end{abstract}


\end{frontmatter}



\section{Introduction}
\label{sec1}

In modern science and engineering design, many practical problems cannot be accurately described by explicit analytical expressions and can only be evaluated through expensive simulations or physical experiments. These problems are commonly categorized as Black-Box Optimization (BBO) problems \cite{11,12,13}. As the complexity and dimensionality of optimization tasks continue to grow, relying solely on direct evaluations of the true objective function becomes prohibitively expensive. To address this challenge, surrogate-assisted evolutionary algorithms (SAEAs) have emerged as an effective class of methods for solving complex optimization problems. By constructing efficient surrogate models to approximate the true objective function, SAEAs enable efficient global search while significantly reducing the number of costly evaluations \cite{14}. This approach has demonstrated remarkable performance across a wide range of domains, including engineering design \cite{15}, hyperparameter tuning \cite{16}, and control system optimization \cite{17,18}. However, although SAEAs have shown outstanding effectiveness in improving optimization efficiency, existing frameworks still rely heavily on rigid, manually crafted components, such as fixed infill criterion design and evolutionary strategy configuration. {This reliance on predefined structures greatly restricts the flexibility of the algorithm, making it difficult to adapt dynamically to task-specific characteristics and the evolving requirements of different search phases \cite{r3}}. Consequently, traditional SAEA methods often lack sufficient transferability and adaptability when faced with diverse optimization tasks and varying search demands \cite{19,20}.

\begin{figure}[h]
    \centering
    \includegraphics[width=0.9\linewidth]{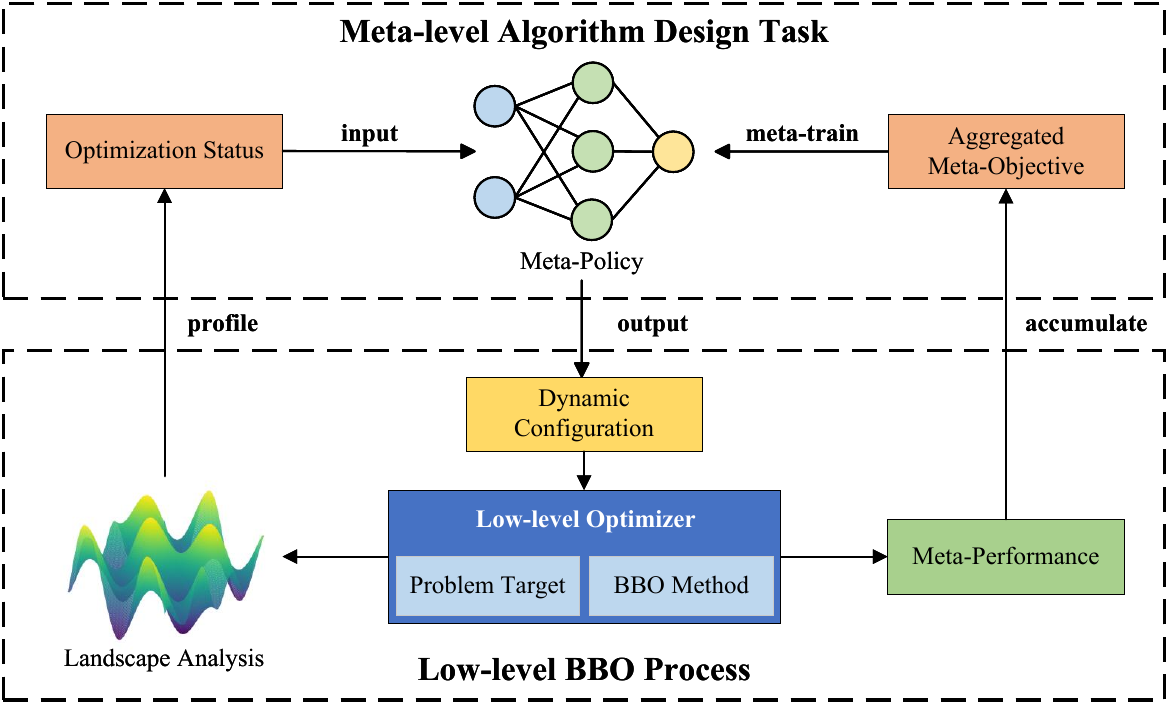}
    \caption{The general workflow of MetaBBO.}
    \label{fig:MetaBBO}
\end{figure}

To overcome the limited flexibility and adaptability of conventional SAEA component design, Meta-Black-Box Optimization (MetaBBO) offers a promising solution. As illustrated in Figure \ref{fig:MetaBBO}, MetaBBO adopts a bi-level optimization framework \cite{13}. {At the meta level, a control policy (often realized through a neural network) adaptively determines algorithm configurations for the lower-level BBO optimizer based on its current search state, which is characterized using Exploratory Landscape Analysis (ELA) \cite{r1}}. Once the configuration is applied, the lower-level optimizer solves the target problem and provides performance feedback to the meta controller. The goal of MetaBBO is to meta-learn a policy that maximizes cumulative performance gains over a distribution of tasks \cite{22,23}. By integrating this meta-level decision process, MetaBBO allows adaptive orchestration of SAEA components, including surrogate models, infill criteria, and evolutionary operators, in a manner that depends on both the task and the optimization phase. This dynamic control significantly enhances the algorithm’s generalization ability, transferability, and robustness, enabling it to sustain efficient search performance across complex and evolving optimization scenarios \cite{24}.

Despite the growing interest in MetaBBO, most existing frameworks still adopt traditional evolutionary algorithms as their low-level optimizers. Existing examples include DE-DDQN \cite{26}, which focuses on mutation operator selection in differential evolution; RLLPSO \cite{27}, which adaptively adjusts the number of performance levels in the particle swarm population; and HHRL-MAR \cite{28}, which dynamically switches swarm intelligence optimizers during the search process via a Q-table-based reinforcement learning agent. In contrast, only a limited number of studies have explored using SAEA as the low-level optimizer, and these works still face substantial limitations. This is mainly because SAEA involves multiple interacting components, making comprehensive and fine-grained control challenging. For instance, DRL-SAEA \cite{29} provides three candidate surrogate modeling schemes but only controls the selection of surrogate models, leaving other crucial components—such as infill criterion and evolutionary strategy—untouched. As a result, the full potential of MetaBBO for joint control of multiple SAEA components remains largely underexplored.

In addition to the aforementioned issues, current research on SAEA rarely considers the trade-off between the robustness and predictive accuracy of surrogate models. Some studies have introduced bagging-based ensemble strategies as surrogate modeling methods \cite{32,31,33}. Owing to their low estimation variance, bagging approaches exhibit strong robustness, making them particularly suitable for evolutionary algorithms based on ranking criteria, where high estimation precision is not essential during the early search stage \cite{30}. However, these methods inherently involve estimation bias, which can significantly affect optimization performance during the convergence phase, especially when precise evaluations are required. In such scenarios, boosting strategies may offer a better fit, as they can reduce bias and provide more accurate surrogate predictions in later optimization stages \cite{34}. 

To promote the integration of SAEAs into the MetaBBO paradigm and advance research on balancing the robustness and accuracy of surrogate models, we propose a novel method named AdaE-SAEA (Adaptive Ensemble Surrogate-Assisted Evolutionary Algorithm) for solving expensive multi-objective optimization problems. AdaE-SAEA incorporates SAEA as the low-level optimizer within the MetaBBO framework and achieves joint control over both the infill criterion and the ensemble-based surrogate modeling strategy. Specifically, two ensemble learning schemes (bagging and boosting) are designed as surrogate modeling modules, together with five infill criteria as candidate sampling strategies, so that the meta-policy can flexibly balance robustness and accuracy, as well as exploration and exploitation, throughout the optimization process. A reinforcement learning approach is employed to train the meta-policy, enabling adaptive decision-making across diverse optimization scenarios. Formally, the optimization problem can be expressed as:

\begin{equation}
    \min_{ \bm{x} \in \Omega} \ \bm{F}(\bm{x}) = 
    \bigl(f_1( \bm{x}), f_2(\bm{x}), \dots, f_m(\bm{x})\bigr),
\end{equation}
where $ \bm{x} = (x_1, x_2, \dots, x_d) $ denotes the $d\text{-dimensional}$ decision vector, $\Omega \subseteq \mathbb{R}^d$ is the search space,  and $f_i(\bm{x})$ represents the $i\text{-th}$ objective function. Since evaluating $\bm{F}(\bm{x})$ typically involves high-fidelity simulations or costly experiments, the number of evaluations is severely limited, which makes traditional optimization strategies inefficient for such problems.

The main contributions of this work can be summarized as follows:

\begin{enumerate}
    \item Ensemble-based surrogate modeling for robustness–accuracy trade-off. We introduce bagging and boosting ensemble learning strategies into SAEA as surrogate modeling approaches. This design allows the optimization process to dynamically adapt to different search phases, enhancing model robustness during early exploration and improving predictive accuracy during late exploitation.

    \item Unified control over modeling strategies and infill criteria based on MetaBBO. Leveraging the MetaBBO framework, we achieve joint control of both the surrogate modeling strategy and the infill criterion, enabling flexible and adaptive configuration of SAEA components. This unified control mechanism significantly improves the algorithm’s transferability and adaptability to different optimization scenarios. In addition, we adopt a distributed sampling and unified training strategy to enhance policy generalization, where the distributed sampling is efficiently implemented using the Ray framework \cite{35}.
    
    \item Efficient Surrogate Modeling with Tabular Prior-Data Fitted Network (TabPFN). To address the inefficiency of conventional surrogate models, e.g., Gaussian Processes (GP), in optimization tasks, we are the first to adopt TabPFN \cite{36} as the base surrogate model for ensemble learning. Unlike traditional GP-based approaches, TabPFN provides fast and reliable predictions, enabling frequent model updates required in MetaBBO.
\end{enumerate}

\section{Related Work}
\label{sec2}

\subsection{Meta-Black-Box Optimization}
\label{subsec2.1}

{
MetaBBO has rapidly become a prominent paradigm for automating the design and configuration of optimization algorithms. Unlike conventional BBO approaches that depend on manually designed algorithms and laborious hyperparameter tuning, MetaBBO leverages meta-level learning to dynamically adapt the settings of the underlying optimizer \cite{22,23,37,38}. This enables optimization strategies to generalize effectively across diverse tasks and adapt to different stages of the search process, significantly reducing the need for expert intervention \cite{d1,d2,d3}. 
}

{MetaBBO typically adopts a bi-level learning framework. The meta-level learns a control policy that determines how to design or configure the optimization process, while the low-level executes the optimization on specific problems and provides performance feedback \cite{13}. {This feedback loop enables the meta-policy to improve over time, achieving adaptive and dynamic control of algorithmic components, such as operators, infill criteria, or model configurations \cite{24,r4}.}}

{Existing MetaBBO methods can be categorized into four main paradigms: 1) algorithm selection, which focuses on identifying the most suitable optimizer from a predefined portfolio for a given task \cite{42,43,44};{ 2) algorithm configuration, which dynamically adjusts hyperparameters and operator settings to improve performance \cite{22,46,47,48,r2};} 3) solution manipulation, where the meta-policy directly controls the generation and evolution of candidate solutions without relying on a fixed optimizer \cite{49}; and 4) algorithm generation, which aims to automatically synthesize new algorithmic workflows or operator compositions \cite{23,51,52}. These tasks are typically addressed using various meta-learning techniques, including reinforcement learning (e.g., policy gradient and actor–critic methods) to learn control policies, supervised learning to approximate task–algorithm mappings, and neuroevolution to directly optimize policy architectures \cite{41}.} {More recently, in-context learning with large language models has emerged as a promising direction, enabling flexible and explainable algorithm design \cite{d5}. Together, these approaches empower MetaBBO to automatically discover effective algorithmic behaviors and adapt strategies across diverse optimization scenarios, reducing reliance on manual design and improving transferability and adaptability \cite{31,37}.}

{Currently, there are relatively few studies on MetaBBO using SAEA as the low-level optimizer, among which the most representative methods are DRL-SAEA and DB-SAEA. DRL-SAEA focuses solely on single-level control of surrogate modeling schemes, whereas DB-SAEA implements dual-level control over SAEA, adjusting both the number of evolutionary algorithm iterations and the infill criterion, and is specifically designed to address multi-objective optimization problems. However, these methods overlook the robustness–accuracy trade-off of surrogate models and do not investigate the integration or ensemble of multiple surrogate models.}

\subsection{Surrogate-Assisted Evolutionary Algorithm for Expensive Optimization Problems}
\label{subsec2.2}
SAEAs have emerged as a powerful class of optimization frameworks specifically designed for expensive optimization problems, where each function evaluation may involve high-fidelity numerical simulations or costly physical experiments \cite{14,17,221}. Such problems are common in fields like aerospace design, structural optimization, energy systems, and process engineering, where reducing the number of expensive evaluations is crucial \cite{222}. SAEAs address this challenge by constructing surrogate models to approximate the objective or constraint functions, thereby guiding the search process more efficiently while preserving the global exploration capability of evolutionary algorithms \cite{223}.

A core component of SAEAs is the surrogate modeling strategy, which replaces or complements direct evaluations with predictive models \cite{224}. Commonly used surrogates include polynomial response surfaces, Kriging or GP models, radial basis functions, support vector machines, and ensemble models \cite{15,16,225}. Each modeling technique has its own advantages in terms of accuracy, uncertainty quantification, and computational complexity, making model selection and management a critical factor for performance.

Beyond surrogate modeling, SAEAs also incorporate infill criteria to determine where to sample next in order to best balance exploration and exploitation \cite{226}. Classical acquisition functions include Expected Improvement (EI), Probability of Improvement (PI), Upper Confidence Bound (UCB), and more sophisticated hybrid criteria tailored for multi-objective and constrained problems \cite{18,227}. These criteria allow SAEAs to adaptively focus computational resources on promising regions of the search space.

SAEAs have been applied to a wide range of optimization scenarios, including single-objective, multi-objective, constrained, and multi-fidelity optimization problems \cite{228}. In multi-objective cases, the surrogate models can either be trained independently for each objective or jointly through vector-valued modeling \cite{229}. In constrained optimization, surrogates are often employed to approximate constraint functions, enabling efficient feasibility estimation and constraint handling. Hybrid frameworks have also been proposed, combining surrogate-assisted techniques with decomposition-based or indicator-based evolutionary algorithms to further enhance scalability and solution diversity \cite{22-10}.

Despite the remarkable progress, several challenges remain. First, surrogate modeling becomes increasingly difficult in high-dimensional, noisy, or multi-objective problems, where prediction accuracy and uncertainty quantification can degrade. Second, efficiently managing computational resources when objective and constraint evaluation costs are heterogeneous remains an open issue. Finally, developing scalable and interpretable surrogate selection mechanisms is essential to fully exploit the potential of SAEAs in complex real-world problems \cite{17,55,54}.

\section{Ensemble-based Surrogate Modeling Strategy}
\label{sec3}

\begin{figure}[h]
    \centering
    \includegraphics[width=0.8\linewidth]{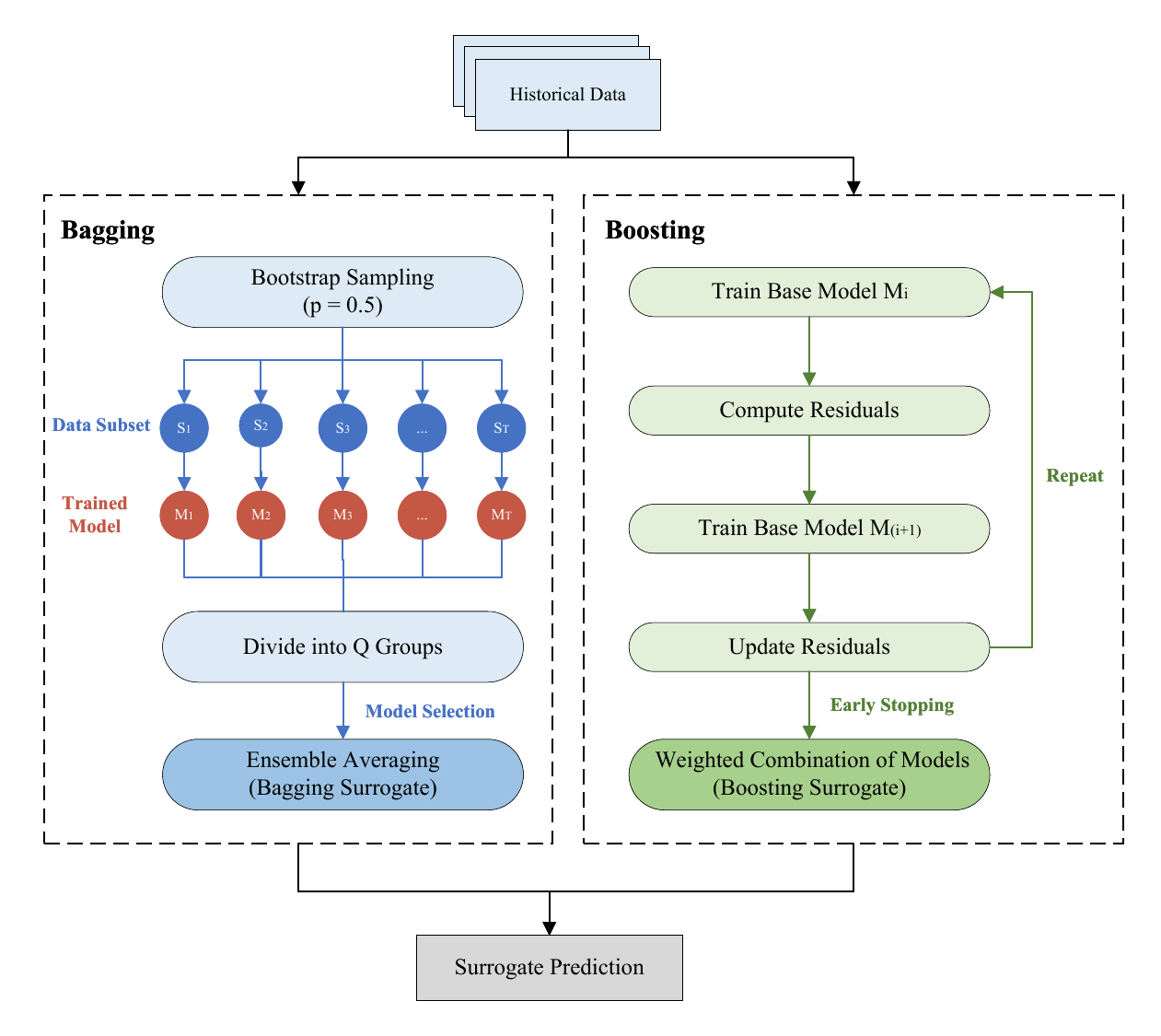}
    \caption{Bagging and boosting surrogate modeling processes.}
    \label{fig:surrogate}
\end{figure}

This section introduces the ensemble modeling strategy used in the AdaE-SAEA method, including bagging and boosting approaches, the use of TabPFN as the base model, and the uncertainty estimation method for ensemble models. The bagging and boosting surrogate modeling processes are shown in Figure \ref{fig:surrogate}.

\subsection{Bagging Surrogate Modeling}
\label{subsec3.1}
Bagging is a classical ensemble learning technique that improves model performance by training multiple base learners on different bootstrap samples and aggregating their predictions \cite{57}. However, we take a different approach from the classical Bagging method. To enhance the accuracy and stability of surrogate modeling in multi-objective optimization, independent bagging ensemble surrogate models are constructed for each objective $f_i$ $(i = 1, 2, \dots, m)$ by applying a half-sampling scheme to the full training dataset. Specifically, each sample is included with a probability of 0.5, resulting in $T$ different bootstrap training sets. A surrogate model is trained on each of these $T$ training sets, forming a diverse model pool $\{ M_{i,1}, M_{i,2}, \dots, M_{i,T} \}$ for the $i\text{-th}$ objective.

Subsequently, we compute the prediction errors of all $T$ models on the Pareto front solutions for the $i\text{-th}$ objective and obtain their mean predictive error as a measure of model accuracy. The models are then sorted according to this accuracy metric, and the sorted model pool is evenly divided into $Q$ groups, where $Q$ is the predefined ensemble size. Finally, one model is randomly selected from each group to form the final bagging ensemble surrogate for the $i\text{-th}$ objective.

By repeating this procedure for all $m$ objectives, we obtain $m$ objective-specific ensemble surrogate models. This design enables each objective to be approximated by a specialized ensemble, achieving both model diversity and estimation accuracy while effectively controlling computational cost. Moreover, by evaluating models on the Pareto front, the ensemble selection is guided toward regions most critical to the optimization process. The overall procedure of Bagging Surrogate Modeling is summarized in Algorithm \ref{alg:ensemble}.

\begin{algorithm}[h]
\caption{Bagging-based Ensemble Surrogate Model Construction}
\small
\label{alg:ensemble}
\begin{algorithmic}[1]
\REQUIRE Training dataset $D = \{(\bm{x}_j, y_j)\}_{j=1}^{N}$ \\    Number of base learners $T$ \\ Ensemble subset size $Q$ \\
\ENSURE $m$ ensemble surrogate models $\{\hat{f}_1, \hat{f}_2, \ldots, \hat{f}_m\}$

\STATE Identify the current Pareto front PF from the population
\FOR{each objective $i = 1$ to $m$}
    \STATE Initialize model pool $\mathcal{M}_i = \{\}$
    \FOR{$t = 1$ to $T$}
        \STATE Construct a bootstrap dataset $D_t$ by half-sampling $D$ ($p=0.5$)
        \STATE Train a base surrogate model $M_{i,t}$ on $D_t$
        \STATE Add $M_{i,t}$ to model pool $\mathcal{M}_i$
    \ENDFOR
    \STATE Evaluate prediction error of each model $M_{i,t}$ on PF (objective $i$)
    \STATE Sort models in $\mathcal{M}_i$ by their average estimation error
    \STATE Evenly divide the sorted models into $Q$ groups
    \STATE Randomly select one model from each group
    \STATE Construct ensemble $\hat{f}_i$ for objective $i$ with the selected $Q$ models
\ENDFOR
\end{algorithmic}
\end{algorithm}

\subsection{Boosting Surrogate Modeling}
\label{subsec3.2}
{For each objective $f_i$, this study adopts Gradient Boosting, a classical boosting approach, to better adapt to the accuracy requirements of the optimization process. Gradient Boosting progressively improves model performance through iterative residual fitting, enhancing its approximation capability and stability on complex objective functions \cite{57,58}. In our implementation, the boosting surrogate is constructed as a stage-wise additive ensemble: at each stage, a base learner is trained to fit the residual errors produced by the current ensemble, and the final prediction is obtained by additively aggregating all learned base learners. We use TabPFN as the base regressor at each boosting stage.}

To avoid underfitting or overfitting, the number of base learners $Q$ is determined using early stopping. By employing this well-established technique as the surrogate modeling approach, the optimization process can obtain more accurate objective estimations in critical search stages, thereby supporting more effective optimization.

\subsection{Base Model and Ensemble-Based Uncertainty Estimation}
\label{subsec3.3}
Traditional surrogate models such as GP suffer from high computational cost and poor scalability as the dimensionality increases. In high-dimensional problems, GP often produce unreliable predictions, which limits their ability to support metapolicy learning. Moreover, the high training cost of GP makes them impractical for MetaBBO frameworks, which typically require frequent model updates across many tasks. In this work, we employ ensemble learning as the surrogate modeling strategy. If GP were used as base models, the computational cost would become prohibitively high, making the approach impractical for MetaBBO. Considering both efficiency and scalability, we choose TabPFN as the base model for our ensemble surrogate, leveraging its strong generalization ability and fast inference to meet the frequent update requirements of MetaBBO. 

TabPFN is a transformer-based probabilistic estimator that achieves high prediction accuracy and low-latency one-shot inference without requiring task-specific training \cite{36,59,60,61}, making it highly suitable for MetaBBO. Specifically, TabPFN directly outputs a probability distribution over predefined objective value intervals. Specifically, for a candidate solution $\bm{x} \in \mathbb{R}^d$, TabPFN returns a probability vector:

\begin{equation}
    p = \text{TabPFN}(\bm{x}) = \left[p^1, \dots, p^K\right] \in [0,1]^K,
\end{equation}
where $K$ is the number of predefined intervals and $\sum_{k=1}^K p^{k}(\bm{x}) = 1$. Each $p^{k}$ represents the predicted probability that the objective value of $\bm{x}$ falls within the $k\text{-th}$ interval. To obtain a scalar prediction and its associated uncertainty, we firstly define the midpoint of each bin as $b'_k = (b_{k-1} + b_k)/2$, where $\{b_k\}_{k=0}^{K}$ denotes the edges of each bin. Then, we estimate the predicted value and standard deviation of $\bm{x}$ using the following equations:

\begin{equation}
\mu{(\bm{x})} = \sum_{k=1}^{K} p^k(\bm{x}) \cdot b'_k ,
\end{equation}

\begin{equation}
{\sigma}(\bm{x}) = 
\sqrt{\sum_{k=1}^{K} p^{k}(\bm{x}) 
\left( b'_k - \mu(\bm{x})\right)^2 } .
\end{equation}

After obtaining the point predictions and uncertainty estimates of individual base models, we further consider how to estimate uncertainty at the ensemble level. From a unified perspective, both bagging and boosting can be viewed as constructing an ensemble predictor through a linear combination of multiple base learners. Let $w_q$ be the weight assigned to the $q\text{-th}$ base model, then the ensemble prediction is:

\begin{equation}
\hat{f}(\bm{x}) = \sum_{q=1}^{Q} w_q \mu_q(\bm{x}) .
\end{equation}

The overall uncertainty can be derived using the law of total variance and the variance of linear combinations:

\begin{align}
\hat{\sigma}_\text{ens}^2(\bm{x}) &= 
\sum_{q=1}^{Q} w_q^2 \sigma_q^2(\bm{x})
+ \sum_{q=1}^{Q} w_q \mu_q^2(\bm{x})
- \left( \sum_{q=1}^{Q} w_q \mu_q(\bm{x}) \right)^2 \notag \\
&\quad + 2 \sum_{s<q} w_s w_q \mathrm{Cov}\big(\mu_s(\bm{x}), \mu_q(\bm{x})\big) .
\end{align}

In practice, bagging typically assumes independence among base learners, so the covariance term is often neglected. In contrast, boosting involves strongly correlated base learners. In this work, we estimate the covariance term by randomly sampling 100 points in the neighborhood of the solution $\bm{x}$ and computing the sample covariance of the model predictions.

\section{Adaptive Surrogate Ensemble Optimization}
\label{sec4}
In this section, we introduce how AdaE-SAEA integrates the two ensemble surrogate modeling strategies, to achieve unified control of both the robustness–accuracy trade-off and the infill criterion within SAEA. By formulating the optimization as a Markov Decision Process (MDP), a meta-policy is learned to coordinate modeling strategies and infill criteria adaptively. We then describe the training process of this meta-policy, including state representation, action space design, and reward formulation.

\subsection{Procedure of AdaE-SAEA}
\begin{figure}[h]
    \centering
    \includegraphics[width=0.9\linewidth]{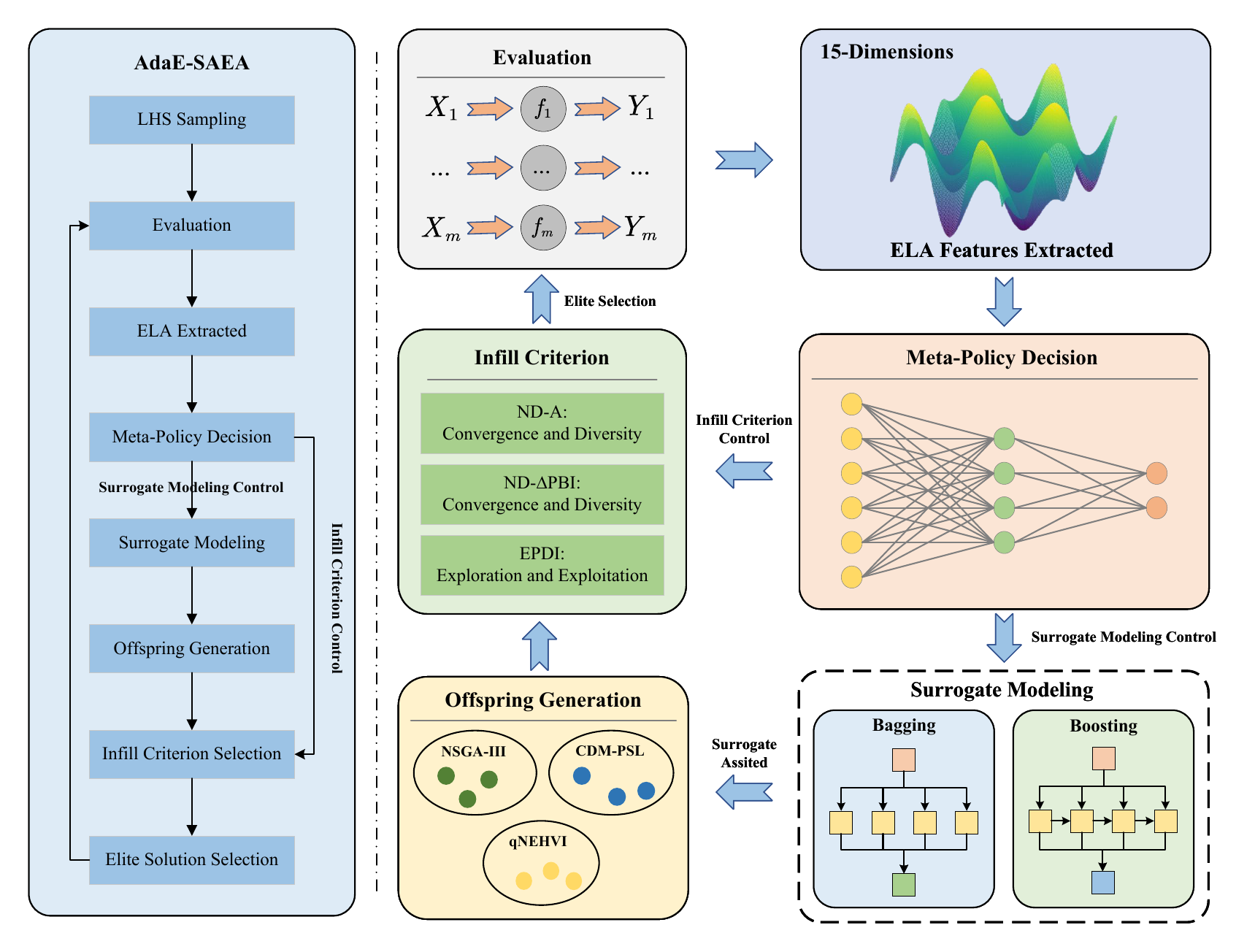}
    \caption{The workflow of AdaE-SAEA.}
    \label{fig:AdaE}
\end{figure}
\label{subsec4.1}

The optimization process of AdaE-SAEA begins with initial sampling conducted through Latin Hypercube Sampling (LHS) to ensure sufficient coverage of the search space, followed by expensive evaluations of the sampled solutions. Before each expensive evaluation, ELA features are extracted from the current population to characterize the optimization state, which is then fed into the meta-policy. The meta-policy adaptively determines both the surrogate modeling strategy (bagging or boosting) and the infill criterion. Based on this configuration, the SAEA optimizer builds surrogate models with historical data. To improve the quality and diversity of search, we design a hybrid candidate generation mechanism that integrates NSGA-III \cite{a31}, CDM-PSL \cite{a32}, and qNEHVI \cite{a33}. Candidate solutions are generated through this hybrid sampling strategy, then refined through several surrogate-assisted evolutionary steps. Finally, the infill criterion identifies the most promising candidates for expensive evaluations, and the newly evaluated data are incorporated into the training set for the next iteration. The overall procedure is summarized in Algorithm \ref{alg:adae-saea} and illustrated in Figure \ref{fig:AdaE} for a clearer overview of the optimization framework.

\begin{algorithm}[htb]
\caption{AdaE-SAEA}
\small
\label{alg:adae-saea}
\begin{algorithmic}[1]
\REQUIRE Multi-objective black-box function $\bm{F}(\cdot)$, evaluation budget $FE_{\text{max}}$, meta-policy $\pi_{\theta}$, initial sample size $n_0$, batch size $n_{bs}$
\ENSURE Expensive evaluated solutions

\STATE $X \leftarrow$ Get $n_0$ initial inputs by LHS
\STATE $Y \leftarrow$ Objective functions evaluation by $\bm{F}(\cdot)$
\STATE Construct initial dataset $D \leftarrow \{X, Y\}$, $t = |D|$
\WHILE{$t < FE_{\text{max}}$}
    \STATE Extract ELA features $\bm{s}_t$ as the optimization state from $D$
    \STATE $\pi_{\theta}(\bm{s}_t)$ outputs surrogate modeling mode and infill criterion
    \STATE Build surrogate model $\hat{\bm{F}}(\cdot)$ using Bagging or Boosting:
    \IF{Bagging}
        \STATE Perform probabilistic bootstrap sampling ($p = 0.5$)
        \STATE Train $T$ TabPFN base models
        \STATE Select $Q$ models and average predictions
    \ENDIF
    \IF{Boosting}
        \STATE Iteratively fit residuals with $Q$ TabPFN base models
        \STATE Combine weighted base learners for final prediction
    \ENDIF
    \STATE $X_{\text{off}} \leftarrow$ Generate offspring by hybrid strategy and surrogate model $\hat{\bm{F}}(\cdot)$
    \STATE $X^* \leftarrow$ Select top $n_{bs}$ candidates based on the infill criterion
    \STATE $Y^* \leftarrow \bm{F}(X^*)$
    \STATE $D \leftarrow D \cup \{X^*, Y^*\}$
    \STATE $t \leftarrow t + n_{bs}$
\ENDWHILE
\end{algorithmic}
\end{algorithm}

\subsection{Low-level Optimizer Based on SAEA}
\label{subsec4.2}
In the proposed AdaE-SAEA framework, the low-level optimizer plays a crucial role in executing the actual search and optimization process under the guidance of the meta-policy. To ensure efficient use of limited evaluation budgets, we employ a SAEA as the low-level optimizer. SAEA integrates surrogate models into the evolutionary optimization process to approximate expensive objective functions, thereby enabling efficient global search with significantly fewer real evaluations. Within this structure, the meta-policy determines the modeling strategy and infill criterion, while the SAEA is responsible for generating candidate solutions, performing surrogate-assisted evolution, and selecting promising points for expensive evaluations by infill criterion. In general, an SAEA framework consists of three key components: (1) surrogate model, (2) evolutionary search algorithm, and (3) infill criterion. Since the surrogate modeling strategies have been elaborated in the previous section, this part mainly focuses on the evolutionary search algorithm and the infill criteria within the AdaE-SAEA framework.

\subsubsection{Evolutionary Algorithm}
\label{subsubsec4.2.1}
In the evolutionary search process, we adopt a hybrid candidate generation mechanism that integrates NSGA-III, CDM-PSL, and qNEHVI. Previous studies have demonstrated the advantages of hybrid sampling strategies in improving search efficiency \cite{a32,a46,a47}. Specifically, by leveraging NSGA-III for broad exploration and combining it with CDM-PSL and qNEHVI for localized exploitation, the hybrid design in AdaE-SAEA effectively enhances the diversity and structural richness of candidate solutions. Under normal circumstances, the three candidate generation strategies are executed independently, producing diverse candidate sets that are subsequently merged into a unified pool for further selection. To prevent search stagnation, we monitor the average hypervolume (HV) \cite{a48} growth rate over two consecutive evaluations. If the growth rate drops below 0.05, an evolutionary refinement loop is activated. In this loop, crossover and mutation operations, combined with NSGA-III environmental selection, are repeatedly applied to the merged pool, thereby reinforcing both global exploration and local convergence. The overall procedure is summarized in Algorithm \ref{alg:candidate-generation}.

\begin{algorithm}[!htb]
\caption{Candidate Generation of AdaE-SAEA}
\small
\label{alg:candidate-generation}
\begin{algorithmic}[1]
\REQUIRE Evaluated population set $D$, surrogate models $\bm{\hat{F}}(\cdot)$, HV history set $\{HV_{t-2}, HV_{t-1}, HV_t\}$
\ENSURE Offspring population set $D_{\text{off}}$

\STATE $X_{\text{NSGA-III}} \leftarrow \text{NSGA-III}(D)$
\STATE $X_{\text{CDM-PSL}} \leftarrow \text{CDM-PSL}(D)$
\STATE $X_{\text{qNEHVI}} \leftarrow \text{qNEHVI}(D)$
\STATE $X_{\text{off}} \leftarrow X_{\text{NSGA-III}} \cup X_{\text{CDM-PSL}} \cup X_{\text{qNEHVI}}$
\STATE Compute recent HV growth rate:
\STATE $r_1 = \frac{HV_t - HV_{t-1}}{HV_{t-1}}, \quad r_2 = \frac{HV_{t-1} - HV_{t-2}}{HV_{t-2}}$
\IF{$(r_1 + r_2)/2 < 0.05$}
    \STATE $X_{\text{mix}} \leftarrow X_{\text{off}}$
    \FOR{$t = 1$ to $T$}
        \STATE $X_{\text{mix}} \leftarrow \text{Reproduction}(X_{\text{mix}}) \cup X_{\text{mix}}$
        \STATE Evaluate $X_{\text{mix}}$ using surrogate model $\bm{\hat{F}}(\cdot)$
        \STATE $X_{\text{mix}} \leftarrow \text{EnvironmentalSelection}(X_{\text{mix}})$
    \ENDFOR
    \STATE $X_{\text{off}} \leftarrow X_{\text{off}} \cup X_{\text{mix}}$
\ENDIF
\STATE $D_{\text{off}} \leftarrow \{X_{\text{off}}, \bm{\hat{F}}(X_{\text{off}})\}$

\end{algorithmic}
\end{algorithm}

\subsubsection{Infill Criteria}
\label{subsubsec4.2.2}
In surrogate-assisted optimization, the infill criterion determines which candidate solutions are selected for expensive evaluations. It provides a principled mechanism to balance exploration and exploitation, guiding the search toward promising regions while maintaining diversity. By prioritizing evaluation points, infill criteria reduce costly evaluations and accelerate convergence.
For the design of infill criteria, our approach is inspired by the EIC-MSSAEA method \cite{a1}. We adopt five different Infill Criteria, including ND-A, two variants of $\text{ND-}\Delta{PBI}$ that respectively highlight convergence and diversity, and two types EPDI criteria targeting exploration and exploitation. The formal definitions of these criteria are presented in \ref{app1}.

\subsection{Formulating the Evolutionary Search as an MDP}
\label{subsec4.3}
The overall training objective of the proposed framework is to maximize the expected performance across tasks \cite{13}. Formally, let $\mathcal{T}_i$ denote a specific optimization task, and $\pi_\theta$ represent the meta-policy parameterized by $\theta$. The training objective can be expressed as:

\begin{equation}
J(\theta) = \mathbb{E}_{\mathcal{T} \sim \mathcal{D}} \big[ R(\mathcal{O}, \pi_\theta, \mathcal{T}) \big]
\approx \frac{1}{N} \sum_{i=1}^N \sum_{t=1}^T \mathrm{perf}(\mathcal{O}, a_t, \mathcal{T}_i),
\label{eq:7}
\end{equation}
where $\mathrm{perf}(\cdot)$ denotes a task-specific performance metric, $N$ is the number of training tasks, and $R(\cdot)$ represents the total accumulated reward along the search trajectory. To achieve this objective, we formulate the surrogate-assisted evolutionary optimization process as a discrete-time, finite-horizon MDP, defined by a tuple $\mathcal{M} = (\mathcal{S}, \mathcal{A}, P, r, \gamma),$ where $\gamma$ is the discount factor.

\subsubsection{State Space}
\label{subsubsec4.3.1}
In the AdaE-SAEA framework, the state space is designed to provide the meta-policy with sufficient structural and temporal information about the current optimization process. At each time step $t$, the state $s_t\in \mathcal{S}$ is constructed by extracting a set of landscape and progress-related features from the evaluated solutions. These features reflect the current search dynamics, including both the structural properties of the problem landscape and the optimization progress with respect to the evaluation budget \cite{a2,a3}.

To ensure that the state representation is both informative and computationally efficient, we select 15-dimensional exploratory landscape analysis features that can effectively characterize the underlying optimization problem. These state features capture essential geometric, statistical, and progress-related properties of the search, enabling the meta-policy to infer the current search phase and problem structure without explicit access to the true objective function. The detailed list of the selected ELA features is provided in \ref{app2}.

\subsubsection{Action Space}
\label{subsubsec4.3.2}
The action space $\mathcal{A}$ in AdaE-SAEA is designed to enable dynamic joint control over the surrogate modeling strategy and the infill criterion during the optimization process. Specifically, we define ten discrete actions by pairing five different infill criteria with two ensemble surrogate modeling strategies (bagging and boosting). Let $\mathcal{C} = \{ c_1, c_2, c_3, c_4, c_5 \}$ denote the set of candidate infill criteria, and $\mathcal{M} = \{ m_1, m_2 \}$ represent the set of surrogate modeling strategies. The complete action space is constructed as the Cartesian product:

\begin{equation}
\mathcal{A} = \mathcal{C} \times \mathcal{M}, \quad |\mathcal{A}| = 10.
\label{eq:8}
\end{equation}

Each action $a_t \in \mathcal{A}$ corresponds to a specific pair of infill criterion and surrogate modeling strategy. This formulation allows the meta-policy to adaptively control the trade-off between robustness and accuracy, while simultaneously selecting appropriate infill criteria according to the current search state.

The state transition function $P(s_{t+1} \mid s_t, a_t)$ describes how the optimization state evolves after applying an action. In the AdaE-SAEA framework, this transition is determined by the interaction between the chosen surrogate modeling strategy, the infill criterion, and the underlying optimization dynamics. Since the true transition probability is unknown in black-box problems, we do not model it explicitly. Instead, the meta-policy learns from sampled transitions $(s_t, a_t, r_t, s_{t+1})$ collected during the optimization process, enabling adaptive decision-making without requiring analytical knowledge of $P$.

\subsubsection{Reward Fuction}
\label{subsubsec4.3.3}
In the AdaE-SAEA framework, the reward function $r_t$ at time step $t$ reflects the improvement of the Pareto front in expensive multi-objective optimization tasks. It measures the actual contribution of newly evaluated solutions at iteration $t$ to the expansion of the front. When the front improves, a positive reward proportional to this contribution is assigned; otherwise, a negative reward penalizes ineffective actions. Formally, the reward function is given by the following:

\begin{align}
r_t =
\begin{cases}
1.0 + \lambda \cdot \displaystyle\sum_{i=1}^{n_{bs}} \frac{d_i}{d^{\,i}_{\text{ref}}}, & \text{if the front is improved}, \\[6pt]
-1.0, & \text{otherwise,}
\end{cases}
\label{eq:9}
\end{align}
where $n_{bs}$ denotes the number of newly evaluated solutions, $d_i$ represents the Manhattan distance to the closest point on the previous Pareto front (before improvement), and $d_{\mathrm{ref}}^{\;i}$ is the Manhattan distance from that closest point to the origin. The ratio $\frac{d_i}{d_{\mathrm{ref}}^{\;i}}$ reflects the relative contribution of each evaluated solution to the expansion of the Pareto front. The scaling factor $\lambda$ controls the magnitude of the positive reward. A negative reward of $-1$ discourages the meta-policy from repeatedly selecting unproductive configurations.

This reward design aligns the training objective of the meta-policy with the underlying optimization goal, i.e., maximizing Pareto front improvement while minimizing unnecessary expensive evaluations. By combining positive reinforcement for meaningful front expansion and penalty for stagnation, the meta-policy learns to adaptively choose surrogate modeling strategies and infill criteria that accelerate convergence and maintain robustness across different search phases.

\subsection{Meta-Policy Training Method}
As formalized in Equation (\ref{eq:7}), our training objective is to maximize the expected cumulative reward obtained through the collaboration between the surrogate-assisted optimizer $\mathcal{O}$ and the meta-policy $\pi_\theta$ across multiple optimization tasks $\mathcal{T}_i \sim \mathcal{D}$. {In particular, we train the meta-policy using a Deep Q-Network (DQN) algorithm \cite{c1}.} To enhance the adaptability and generalization of the learned meta-policy, AdaE-SAEA adopts a parallel sampling and centralized training paradigm, which enables efficient interaction across heterogeneous optimization environments.

Specifically, multiple multi-objective optimization tasks run in parallel and interact with the meta-policy to generate diverse state-action-reward trajectories. These trajectories are aggregated into a centralized replay buffer and used to train the meta-policy with shared parameters. This design facilitates stable and data-efficient learning, as the policy can benefit from heterogeneous experiences while maintaining a unified representation. The sampling-training loop is iterated until the meta-policy converges.

To support parallel sampling, we employ Ray \cite{35}, an open-source framework for distributed and parallel processing. Using Ray, multiple optimization tasks can be executed simultaneously on different CPUs and GPUs, enabling large-scale data collection and efficient meta-policy updates.

\section{Experiment Studies}
\label{sec5}

\subsection{Experimental Setting}
\label{subsec5.1}
To evaluate the performance of AdaE-SAEA, we first set $n_{bs} = 3$ and train the meta-policy across multiple optimization environments. Subsequently, AdaE-SAEA is compared with several state-of-the-art and classical algorithms, including TSEMO \cite{d6}, DGEMO \cite{d7}, CDM-PSL, MOEA/DEGO \cite{a40}, USeMO-EI \cite{a41}, qNEHVI, NSGA-II \cite{a42}, and two MetaBBO-based methods, LRMODE \cite{a43} and R2-RLMOEA \cite{a44}, on ZDT1-3 \cite{a45}, DTLZ2-7 \cite{b2}, and seven real-world benchmark problems \cite{b3}. Moreover, we conduct a sensitivity analysis on $n_{bs}$ under expensive evaluation conditions to investigate its impact on performance. In addition, we perform ablation studies on the evolutionary strategy and meta-policy control mechanism. Finally, we validate the effectiveness of TabPFN as the base surrogate model. This work assesses the performance of the algorithm using the hypervolume (HV) metric to assess the quality of the solutions obtained. The reference points used for HV calculation are aligned with CDM-PSL method and are provided in Table \ref{tab:results1}.

\begin{table}[tb]
    \centering
    \caption{Reference points used for HV calculation.}
    \small
    \label{tab:results1}
    \begin{tabular}{lc}
        \toprule
        Problem & Reference Point \\
        \midrule
        ZDT1 & (0.9901, 5.8380) \\
        ZDT2 & (0.9994, 6.8960) \\
        ZDT3 & (0.9994, 6.0571) \\
        DTLZ2 & (2.8390, 2.9011, 2.8575) \\
        DTLZ3 & (2421.6427, 1905.2767, 2532.9691) \\
        DTLZ4 & (3.2675, 2.6443, 2.4263) \\
        DTLZ5 & (2.6672, 2.8009, 2.8575) \\
        DTLZ6 & (16.8258, 16.9194, 17.7646) \\
        DTLZ7 & (0.9984, 0.9961, 22.8114) \\
        RE1 & (2763.2229, 0.0369) \\
        RE2 & (528107.1899, 1279320.8107) \\
        RE3 & (7.6853, 7.2861, 21.5010) \\
        RE4 & (6.7921, 60.0, 0.48) \\
        RE5 & (0.8745, 1.05092, 1.0533) \\
        RE6 & (749.9241, 2229.3748) \\
        RE7 & (210.3363, 1069.9160, 39196770.1704) \\
        \bottomrule
    \end{tabular}
\end{table}

We employ a parallel sampling strategy to interact with multiple benchmark environments simultaneously. All collected experiences are stored in a centralized replay buffer for meta-policy training. The buffer is cleared and refreshed after each training round to ensure stable learning dynamics. {The meta-policy is parameterized by a four-layer multilayer perceptron (MLP) with layer sizes $15\times64\times32\times10$, where ReLU is used as the activation function for the hidden layers.} The training batch size is set to 64, and the Adam optimizer is used with an initial learning rate of 0.0001. {The scaling factor $\lambda$ is set to 1.} In the bagging-based surrogate modeling, we set $T=20$, $Q=5$.

To evaluate the transferability of the learned meta-policy, we adopt a leave-one-task-out cross-validation strategy across 16 optimization environments. Specifically, in each run, the meta-policy is trained on 15 tasks and tested on the remaining unseen task, which has a higher problem dimensionality. This process is repeated 16 times so that each task is used once as the test task. Moreover, each test round is independently repeated 10 times to ensure statistical reliability. During both training and testing phases, the population is initialized with 80 solutions, and the remaining evaluation budget is set to 80 for the optimization process. Additional configuration details for the training and testing phases will be provided in the subsequent experimental sections.

All experiments are conducted on a compute cluster equipped with 8 × 48 GB VGPUs and 8 × 25-core Intel Xeon Platinum 8481C processors. Leveraging the Ray framework, we achieve efficient task-level parallel sampling and centralized training, reducing each sampling–training cycle to approximately 50 minutes.

\subsection{Training Stability and Convergence}
\label{subsec5.2}

\begin{figure}[h]
    \centering
    \includegraphics[width=0.8\linewidth]{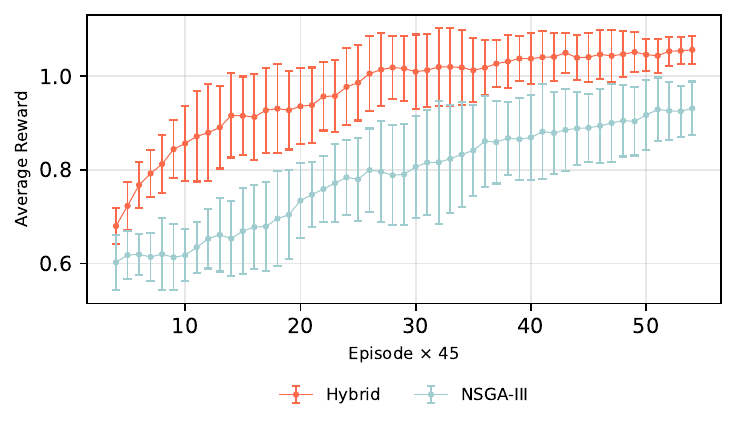}
    \caption{Average reward per expensive evaluation.}
    \label{fig:avg_reward}
\end{figure}

We conduct 16 training runs using a leave-one-task-out cross-validation strategy to evaluate the impact of different candidate generation mechanisms on meta-policy training. In each run, 15 optimization environments are selected for training, and each environment is configured with three dimensional settings (15, 20, and 25), leading to a total of 45 distinct training tasks. Figure \ref{fig:avg_reward} illustrates the average reward per expensive evaluation across all training environments. To improve readability, the curve is smoothed using a moving average with a window size of 5. Vertical bars indicate ±1 standard deviation over 16 runs.

To assess the effectiveness of the hybrid candidate generation mechanism, we train two types of AdaE-SAEA: 1) one equipped with the hybrid candidate generation strategy (NSGA-III $+$ CDM-PSL $+$ qNEHVI), and 2) another using NSGA-III alone as the baseline strategy. As shown in Figure 4, the hybrid strategy consistently achieves higher average rewards and lower standard deviations during training compared to the NSGA-III-only variant. This indicates that incorporating multiple candidate generation strategies enhances search diversity and improves the model’s ability to identify high-quality solutions. Moreover, the reduction in reward standard deviations demonstrates that the hybrid strategy contributes to greater stability in meta-policy training. In subsequent sections, we will further validate this conclusion through experiments on testing problems.

\subsection{Comparisons with Peers}
\label{subsec5.3}
In this section, we evaluate AdaE-SAEA on unseen tasks with higher dimensionality. After training on 15 benchmark tasks, the model is directly tested on a held-out task with 30 decision variables. We compare AdaE-SAEA with five state-of-the-art methods and two MetaBBO-based methods on the ZDT1-3, DTLZ2-7, and seven real-world benchmark problems to evaluate the performance of the proposed algorithm. All algorithmic parameters are configured according to their original papers for fair comparison. {Among the two MetaBBO baselines, LRMODE dynamically controls the selection of evolutionary operators, while R2-RLMOEA uses an MLP-based RL agent to dynamically switch among five EA optimizers during the optimization process.} The MetaBBO-based methods are trained using the same strategy as AdaE-SAEA, i.e., parallel sampling combined with centralized training. The Adam optimizer with the same learning rate is applied uniformly across all methods. The HV results for all algorithms with $n_{bs}=3$ are summarized in Table \ref{tab:2}. Furthermore, the Wilcoxon Signed-Rank Test with a significance level of 0.05 is performed to assess the statistical significance of the results. In the table, the symbols '$+$', '$-$', and '$=$' respectively indicate that the corresponding baseline method performs significantly better, worse, or comparably to AdaE-SAEA.

As shown in Table \ref{tab:2}, AdaE-SAEA achieves the best performance across all ZDT optimization scenarios and outperforms most competing methods on DTLZ and RE problems. Both CDM-PSL and qNEHVI consistently perform better than NSGA-II, MOEA/DEGO, USeMO-EI, TSEMO, DGEMO and the MetaBBO-based methods LRMODE and R2-RLMOEA in most cases, demonstrating the strong potential of Bayesian optimization in multi-objective optimization tasks. Among the MetaBBO-based methods, R2-RLMOEA exhibits the weakest overall performance, which is likely because its design is not specifically tailored for expensive evaluation scenarios.

\begin{table}[h]
  \centering
  \caption{{Comparison of HV performance on unseen 30D tasks.}}
  \label{tab:2}
  \resizebox{\textwidth}{!}{
    \renewcommand{\arraystretch}{2}
    \begin{tabular}{lcccccccccc}
        \toprule
            Problem & TSEMO & DGEMO & CDM-PSL & MOEA/DEGO & USeMO-EI & qNEHVI & NSGA-II & LRMODE & R2-RLMOEA & AdaE-SAEA \\
            \midrule
            ZDT1 & 4.63e+0(1.01e-1)- & 3.45e+0(8.53e-2)- & 5.67e+0(1.37e-1)- & 4.40e+0(1.09e-1)- & 5.17e+0(9.01e-2)- & 5.20e+0(9.97e-2)- & 3.32e+0(8.03e-1)- & 4.69e+0(2.93e-1)- & 3.02e+0(2.46e-1)- & \textbf{5.91e+0(1.91e-1)} \\
            ZDT2 & 5.52e+0(4.23e-2)- & 3.30e+0(1.19e-2)- & 6.12e+0(2.58e-1)- & 5.21e+0(9.99e-2)- & 5.62e+0(1.77e-1)- & 5.33e+0(1.13e-1)- & 3.79e+0(8.03e-2)- & 6.08e+0(6.03e-2)- & 3.78e+0(1.13e-1) & \textbf{6.20e+0(1.83e-1)} \\
            ZDT3 & 5.57e+1(3.87e-1)- & 4.02e+0(2.33e-1)- & 6.03e+0(3.71e-2)= & 4.82e+0(1.03e-1)- & 5.22e+0(7.93e-2)- & 6.02e+0(3.29e-2)- & 3.78e+0(7.93e-2)- & 5.52e+0(3.00e-2)- & 4.00e+0(4.57e-2)- & \textbf{6.03e+0(2.70e-2)} \\
            DTLZ2 & 1.61e+1(9.95e-1)- & 1.99e+1(1.13e+0) & \textbf{2.13e+1(1.04e+0)+} & 1.58e+1(2.11e-1)- & 1.54e+1(2.41e-1)- & 1.83e+1(6.4e-1)- & 1.69e+1(6.13e-1)- & 1.99e+1(3.44e-1)- & 2.09e+1(3.13e-1)- & 2.06e+1(9.24e-1) \\
            DTLZ3 & 1.09e+10(3.22e+8)- & 1.21e+10(2.03e+8)+ & 1.13e+10(5.01e+8)- & 1.22e+10(8.08e+8)+ & 1.12e+10(4.83e+8)+ & \textbf{1.23e+10(1.10e+9)+} & 1.08e+10(2.12e+8)- & 1.12e+10(4.99e+8)+ & 1.03e+10(3.99e+8)- & 1.11e+10(7.91e+8) \\
            DTLZ4 & 5.19e+0(1.02e-1)- & 1.23e+1(9.03e-1)- & 1.50e+1(1.02e+0)- & 5.55e+0(1.33e-1)- & 5.69e+0(2.02e-1)- & 1.11e+1(1.30e+0)- & 5.47e+0(1.44e-1)- & 1.07e+1(1.01e+0)- & 5.70e+0(1.66e-1)- & \textbf{1.68e+1(9.21e-1)} \\
            DTLZ5 & 1.21e+1(9.54e-1)- & 1.34e+1(1.02e+0)- & 1.64e+1(1.44e+0)- & 1.26e+1(2.49e-1)- & 1.17e+1(2.33e-1)- & 1.55e+1(1.32e+0)- & 1.29e+1(2.22e-1)- & 1.42e+1(5.22e-1)- & 1.37e+1(4.10e-1)- & \textbf{1.64e+1(9.9e-1)} \\
            DTLZ6 & 2.41e+3(2.01e+2)- & 4.30e+3(3.37e+2)- & 4.86e+3(2.52e+2)- & 3.88e+3(3.56e+2)- & 3.24e+3(1.22e+2)- & 2.01e+3(2.45e+1)- & 1.81e+3(5.05e+1)- & 3.00e+3(3.77e+2) & 1.99e+3(4.25e+1)- & \textbf{5.02e+3(4.29e+2)} \\
            DTLZ7 & 1.81e+1(8.38e-1)- & 1.37e+1(7.46e-1)- & 1.74e+1(9.07e-1)- & 9.55e+0(1.01e-1)- & 1.68e+1(4.47e-1)- & 1.85e+1(0.55e+0)= & 9.73e+0(1.19e-1)- & 1.85e+1(1.00e+0)= & 1.80e+1(1.00e+0)- & \textbf{1.85e+1(1.02e+0)} \\
            RE1 & 2.93e+1(6.03e-1)- & 2.75e+1(1.01e+0)- & \textbf{3.16e+1(5.66e-1)+} & 2.70e+1(2.21e-1)- & 2.86e+1(5.01e-1)- & 2.71e+1(4.01e-1)- & 2.78e+1(5.10e-1)- & 3.00e+1(5.15e-1)= & 2.88e+1(4.49e-1)- & 3.00e+1(5.66e-1) \\
            RE2 & 6.73e+11(1.00e+9)- & 6.70e+11(9.10e+8)- & 6.74e+11(1.10e+9)- & 6.74e+11(8.99e+9)- & 6.74e+11(1.14e+9)- & 6.71e+11(1.01e+9)- & 6.72e+11(1.07e+9)- & 6.72e+11(9.77e+8)- & 6.72e+11(1.23e+9)- & \textbf{6.75e+11(1.57e+9)} \\
            RE3 & 8.70e+2(9.02e+0)- & 8.66e+2(9.41e+0)- & 8.87e+2(8.44e+0)- & 8.56e+2(9.21e+0)- & 8.78e+2(8.92e+0)- & 8.90e+2(7.89e+0)- & 8.91e+2(9.85e+0)- & 8.89e+2(1.10e+1)- & 8.90e+2(1.34e+1)- & \textbf{8.98e+2(7.43e+0)} \\
            RE4 & 1.11e+2(5.12e+0)- & 9.53e+1(2.29e+0)- & 1.18e+2(3.89e+0)- & 6.83e+1(3.34e+0)- & 9.78e+1(3.18e+0)- & 1.16e+2(2.22e+0)- & 8.18e+1(2.79e+0)- & 1.18e+2(3.92e+0)- & 9.00e+1(4.90e+0)- & \textbf{1.20e+2(4.01e+0)} \\
            RE5 & 7.20e-1(4.34e-3)- & 7.15e-1(4.88e-3)- & 7.33e-1(3.03e-3)- & 6.62e-1(4.01e-3)- & 6.71e-1(3.24e-3)- & \textbf{7.45e-1(1.01e-2)+} & 5.35e-1(5.21e-3)- & 6.31e-1(5.24e-3)- & 6.24e-1(5.01e-3)- & 7.39e-1(2.99e-3) \\
            RE6 & 1.61e+6(1.38e+4)- & 1.59e+6(9.08e+3)- & 1.62e+6(8.21e+3)- & 1.61e+6(6.98e+3)- & 1.61e+6(7.70e+3)- & 1.62e+6(7.91e+3)- & 1.59e+6(1.11e+4)- & 1.61e+6(7.17e+3)- & 1.61e+6(9.06e+3)- & \textbf{1.63e+6(7.60e+3)} \\
            RE7 & 8.81e+12(7.73e+9)= & 8.79e+12(9.85e+9)- & 8.81e+12(4.93e+9)= & 8.80e+12(6.91e+9)- & 8.80e+12(9.88e+9)- & 8.80e+12(9.89e+9)- & 8.79e+12(1.02e+10)- & 8.80e+12(5.85e+9)- & 8.79e+12(5.50e+9)- & \textbf{8.81e+12(5.49e+9)} \\
            (+/=/-) & (0,1,15) & (1,0,15) & (2,2,12) & (1,0,15) & (1,0,15) & (2,1,13) & (0,0,16) & (1,2,13) & (0,0,16) & \\
            \bottomrule
    \end{tabular}
    }
\end{table}

\begin{figure}[htb]
  \centering
  \subcaptionbox{sensitivity analysis of $n_{bs}$\label{fig:5a}}[0.32\textwidth]{
    \includegraphics[width=\linewidth]{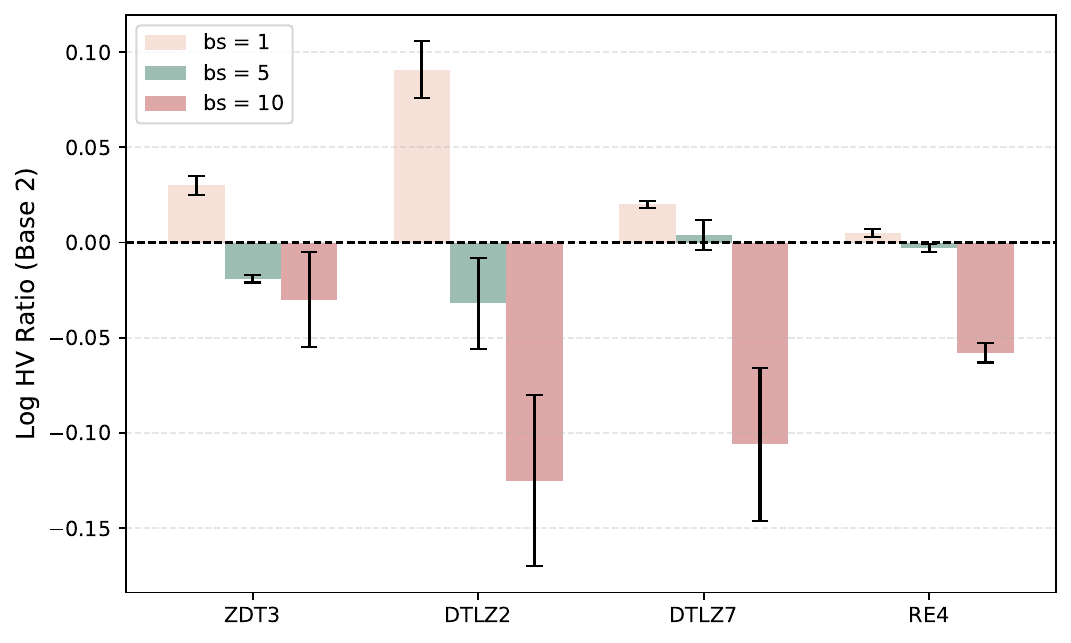}}
  \hfill
  \subcaptionbox{ablation of hybrid candidate generation strategy\label{fig:5b}}[0.32\textwidth]{
    \includegraphics[width=\linewidth]{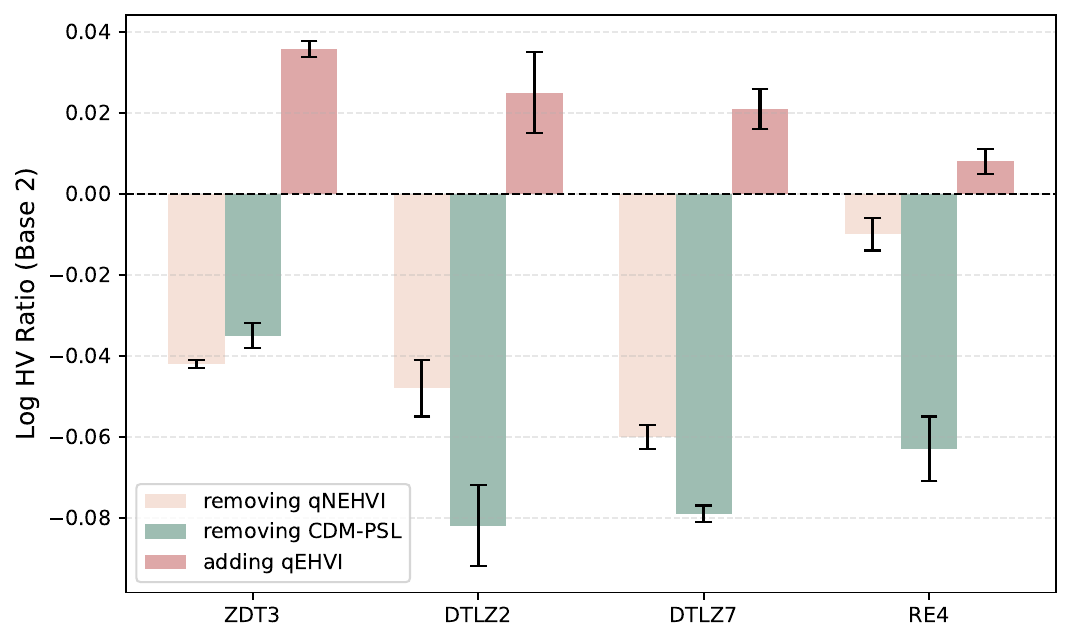}}
  \hfill
  \subcaptionbox{ablation of unified control mechanism\label{fig:5c}}[0.32\textwidth]{
    \includegraphics[width=\linewidth]{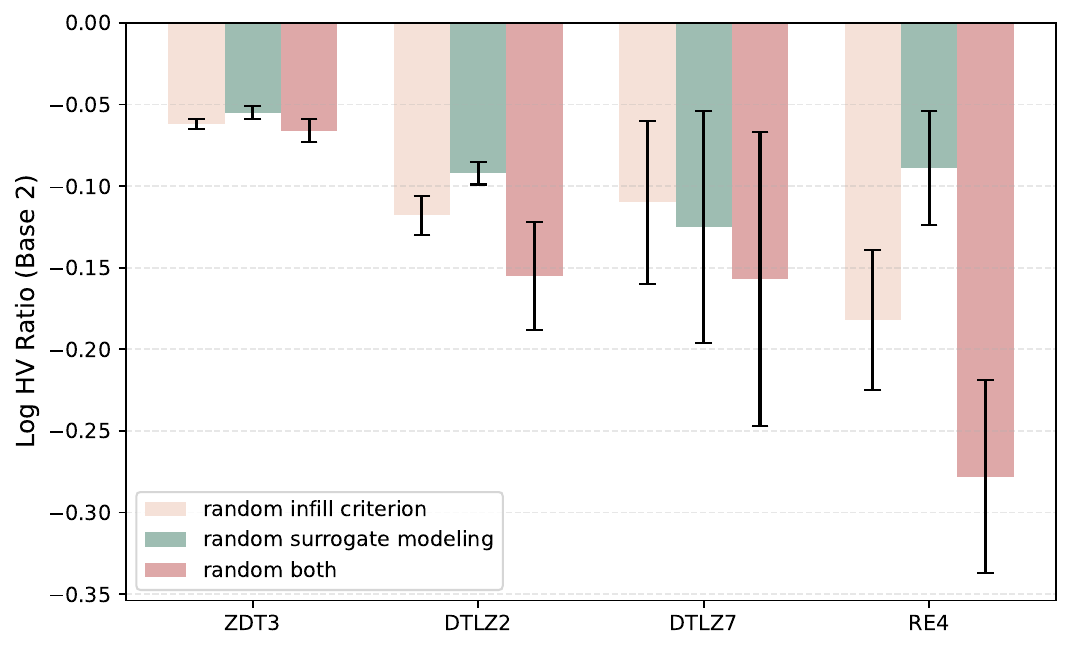}}

  \caption{\scriptsize Sensitivity analysis and ablation study of AdaE-SAEA using logarithmic HV ratio (base 2)}
  \label{fig:comparison}
\end{figure}

\subsection{Sensitivity Analysis of $n_{bs}$}
\label{subsec5.4}

To examine the sensitivity of the proposed meta-policy to different values of $n_{bs}$, we evaluate the performance of a meta-policy trained with $n_{bs}=3$ under alternative settings of $n_{bs}=1,5$ and 10  during the testing phase. This analysis aims to investigate how varying the number of evaluated solutions per iteration influences the overall optimization performance. Figure \ref{fig:5a} presents the logarithmic hypervolume ratio comparing $n_{bs}=1,5$ and 10 against the baseline setting of $n_{bs}=3$ across several randomly selected benchmark problems, including ZDT3, DTLZ2, DTLZ7, and RE4.

As shown in Figure \ref{fig:5a}, slight adjustments to $n_{bs}$ result in only marginal differences in optimization performance compared to the baseline setting of $n_{bs}=3$, and in some cases, $n_{bs}=1$ even achieves slightly better results. {This may be because smaller $n_{bs}$ provides more frequent surrogate updates under limited budgets, reducing the accumulation of surrogate bias and occasionally leading to slightly better performance.} However, when $n_{bs}$ deviates excessively from the baseline, the optimization performance degrades significantly and exhibits larger fluctuations. This observation indicates that overly large $n_{bs}$ values may introduce instability and reduce the efficiency of the search process. Overall, these results suggest that AdaE-SAEA maintains strong robustness with respect to different $n_{bs}$ settings, showing stable optimization performance under moderate variations. This property makes the proposed method more practical and reliable in real-world scenarios where the evaluation batch size may vary.

\subsection{Ablation Study}
\label{subsec5.5}
This section investigates the effectiveness of two key components in AdaE-SAEA: the hybrid candidate generation strategy in the evolutionary algorithm and the unified control mechanism of the meta-policy.

For the hybrid candidate generation strategy, we compare three variants by selectively removing or modifying specific generation components: 1) removing CDM-PSL, 2) removing qNEHVI, and 3) adding qEHVI \cite{a33} as an additional strategy. This comparison aims to analyze the necessity and contribution of each component to the overall optimization performance. As shown in Figure \ref{fig:5b}, incorporating an additional generation strategy slightly improves performance, indicating that increased sampling diversity provides some benefit. Removing qNEHVI leads to a minor decline in performance, while removing CDM-PSL causes a more significant drop, highlighting its stronger contribution to the hybrid mechanism.

For the ablation study on the unified control mechanism, we compare three variants of AdaE-SAEA: 1) controlling only the infill criterion while randomly selecting between bagging and boosting for surrogate modeling, 2)controlling only the surrogate modeling strategy while randomly selecting the infill criterion, and 3)randomly selecting both surrogate modeling and infill criterion. As shown in Figure \ref{fig:5c}, all three variants perform noticeably worse than the original AdaE-SAEA. Notably, the variant that removes the control over surrogate modeling exhibits a more significant performance drop compared to the one that removes control over the infill criterion, indicating that surrogate modeling plays a more critical role in performance. The worst results are obtained when both controls are removed, which aligns with expectations since no adaptive guidance is provided to the optimization process.

\subsection{Effect of TabPFN}
\label{subsec5.6}

\begin{figure}[h]
  \centering
  \subcaptionbox{ZDT1\label{fig:6a}}[0.32\textwidth]{
    \includegraphics[width=\linewidth]{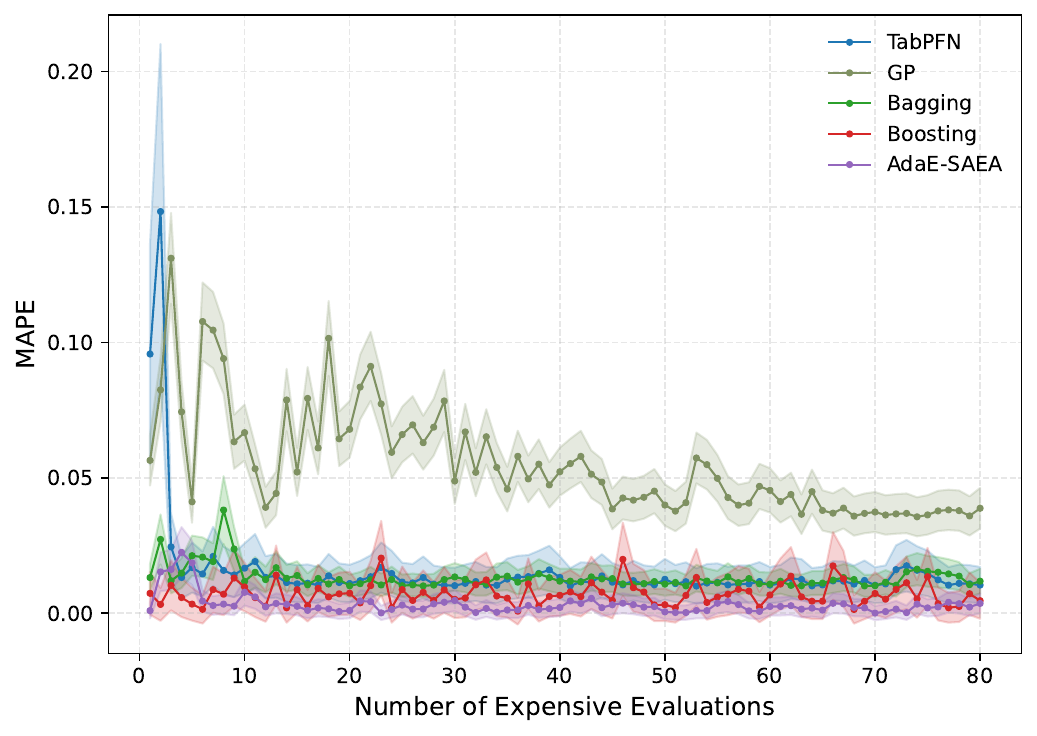}}
  \hfill
  \subcaptionbox{DTLZ4\label{fig:6b}}[0.32\textwidth]{
    \includegraphics[width=\linewidth]{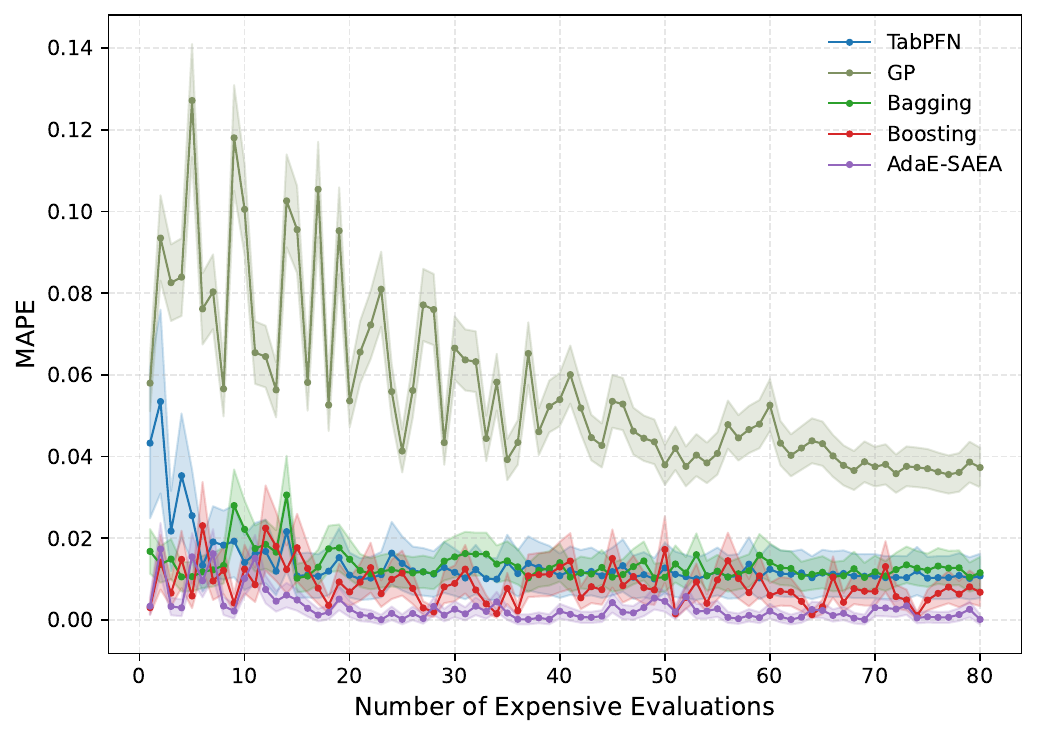}}
  \hfill
  \subcaptionbox{RE5\label{fig:6c}}[0.32\textwidth]{
    \includegraphics[width=\linewidth]{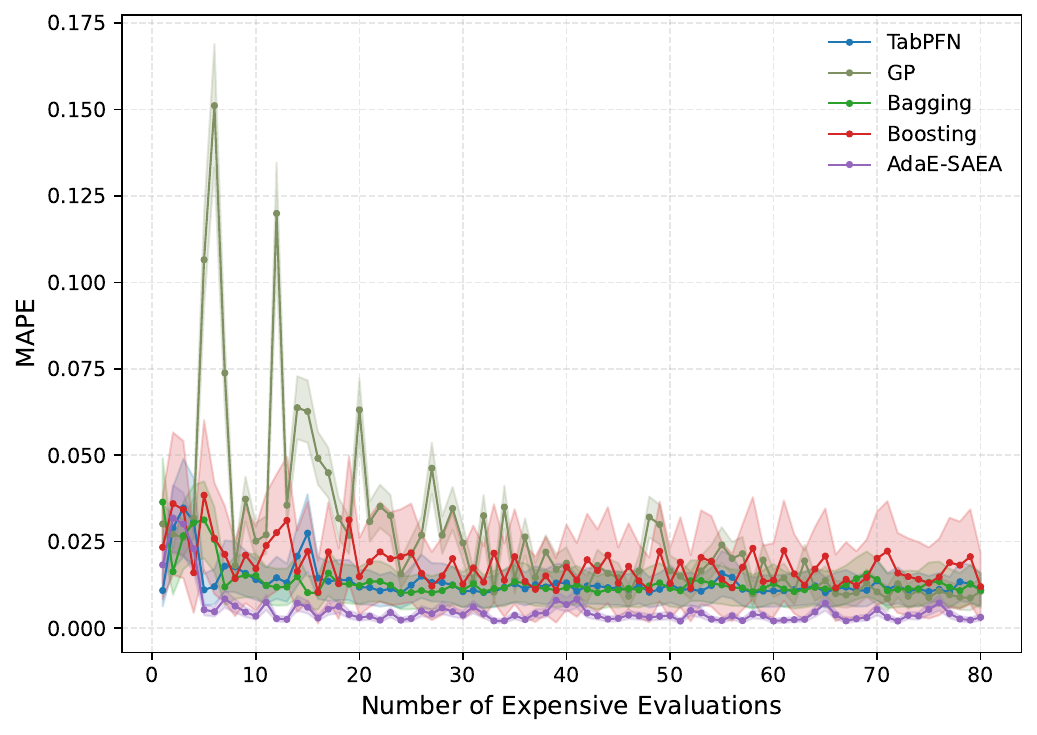}}

  \caption{\scriptsize Comparison of surrogate modeling strategies}
  \label{fig:comparison2}
\end{figure}

This section investigates the effectiveness of TabPFN as a surrogate model in MetaBBO for expensive black-box optimization tasks and its capability as a base model for ensemble learning. We conduct experiments on four representative optimization environments: ZDT1, DTLZ4, and FE5. In these experiments, we use AdaE-SAEA with $n_{bs}=1$, where one candidate solution is selected for expensive evaluation at each iteration. Five modeling strategies are compared: 1) TabPFN as the surrogate model, 2) GP as the surrogate model, 3) Bagging ensemble, 4) Boosting ensemble, and 5) AdaE-SAEA adaptive modeling strategy. Figure \ref{fig:comparison2} illustrates the mean MAPE between predicted and true objective values throughout the expensive evaluation process, with shaded areas indicating the model uncertainty estimates. The MAPE calculation is provided in Equation(\ref{eq:10}). Since this work focuses on multi-objective optimization, we predict each objective function separately and compute the MAPE and uncertainty for each objective. The final values reported in the figure are obtained by averaging the MAPE and uncertainty across all objectives.

\begin{align}
\label{eq:10}
\text{MAPE} = \frac{100\%}{n} \sum_{i=1}^{n} \left| \frac{y_i - \hat{y}_i}{y_i+0.01} \right|.
\end{align}

As shown in the Figure \ref{fig:comparison2}, TabPFN exhibits more accurate approximation performance than GP when used as a surrogate model. Bagging demonstrates stronger robustness than both TabPFN and Boosting, particularly during the early stages of optimization. Although Boosting can sometimes provide more accurate predictions, it suffers from larger fluctuations. In contrast, the meta-policy–controlled surrogate modeling strategy in AdaE-SAEA consistently achieves both high robustness and high accuracy, demonstrating a clear overall advantage.

\section{Conclusion}
\label{sec6}

In this paper, we propose AdaE-SAEA, a MetaBBO framework designed for expensive multi-objective optimization tasks. The proposed method learns a meta-policy to jointly control surrogate modeling strategies and infill criterion selection, enabling adaptive configuration across different optimization stages. By introducing bagging and boosting as ensemble surrogate modeling strategies, AdaE-SAEA effectively addresses the robustness–accuracy trade-off, ensuring both stable early-stage exploration and precise late-stage exploitation. Experimental results demonstrate that AdaE-SAEA achieves strong and robust performance across various benchmark problems and shows low sensitivity to the parameter $n_{bs}$. Furthermore, we validate the effectiveness of TabPFN as the base surrogate model for ensemble learning, highlighting its efficiency in MetaBBO training and its ability to balance robustness and accuracy. Future work will focus on three directions. First, the current joint control mechanism is implemented through a simple Cartesian product, which increases the search space and reduces efficiency. More advanced reinforcement learning approaches, such as hierarchical RL, could improve scalability. Second, while ensemble learning provides robust and accurate modeling, it is computationally expensive, and more efficient modeling strategies are needed. Finally, we plan to extend AdaE-SAEA to more challenging scenarios, including constrained and dynamic multi-objective optimization problems.

\appendix
\section{Infill Criteria of AdaE-SAEA}
\label{app1}

In the AdaE-SAEA framework, we design five types of infill criteria to guide the selection of solutions for expensive evaluations. Specifically, these criteria include ND-A, two variants of $\text{ND-}\Delta \text{PBI}$ that emphasize convergence and diversity, and two EPDI criteria focusing on exploration and exploitation \cite{a1}. The mathematical definitions and implementation details of these criteria are provided below.

ND-A is an angle-based infill criterion designed to promote both convergence and diversity in multi-objective optimization. Given the candidate set $P^*$ and the archive $A_r$, non-dominated sorting is first applied to obtain the non-dominated subsets $P^*_{\text{nd}}$ and $A_{\text{rnd}}$. For $\bm{x} \in P^*_{\text{nd}}$ and $y \in A_{\text{rnd}}$, the minimum angle $\theta^{\min}_x$ is calculated as:

\begin{equation}
\theta_{xy} = \arccos \left( \frac{ \sum_{i=1}^m \hat{f}_i(\bm{x}) \cdot {f}_i(y) }{ \sqrt{ \sum_{i=1}^m \hat{f}_i(\bm{x})^2 } \cdot \sqrt{ \sum_{i=1}^m {f}_i(y)^2 } } \right),
\end{equation}

\begin{equation}
\theta^{\min}_{\bm{x}} = \min_{\bm{x} \in P^*_{\text{nd}},y \in A_{\text{rnd}}} \theta_{xy},
\end{equation}
where $f_i(\cdot)$ denotes the true value of the $i\text{-th}$ objective function, and $\hat{f}_i(\cdot)$ denotes the surrogate regression function trained to approximate the $i\text{-th}$ objective. The candidate with the largest $\theta^{\min}_x$ is then selected, as it represents the solution with the most distinct objective direction relative to the current archive. This encourages the discovery of new search directions and improves the spread of solutions on the Pareto front.

$\text{ND-}\Delta PBI$ is the second criterion, designed to simultaneously consider convergence and diversity, based on the PBI (penalty-based boundary intersection) method. It evaluates the improvement of each candidate solution with respect to the closest nonempty reference vector in the objective space. The PBI value of a solution $\bm{x}$ with respect to reference vector $V_k$ is computed as:

\begin{equation}
 d_{k,1}(\bm{x}) =  \frac{\left| \vec{F}(\bm{x}) \cdot V_k \right|}{ \left| V_k  \right| } ,
 \end{equation}

 \begin{equation}
 d_{k,2}(\bm{x}) = \left| \vec{F}(\bm{x}) - d_{k,1}(\bm{x}) \cdot \frac{V_k}{ \left| V_k \right| } \right|,
 \end{equation}

\begin{equation}
 PBI_k(\bm{x}) = d_{k,1}(\bm{x}) + \lambda \cdot d_{k,2}(\bm{x}),
 \end{equation}
 where $\vec{F}(x)$ is the normalized objective vector, and $d_{k,1}(x)$, $d_{k,2}(x)$ represent the distance from $\vec{F}(x)$ to the reference direction and its perpendicular deviation, respectively. The penalty coefficient $\lambda$ balances convergence and diversity. The solution’s improvement is then measured by the reduction in its projected PBI value:

\begin{equation}
 \Delta PBI_j(\bm{x}) = PBI_{j,\min} - \hat{PBI}_j(\bm{x}), \quad j \in \{{1, ..., |V'_n|}\},
\end{equation}
where $PBI_{j,\min}$ denotes the minimum PBI value among archive solutions along reference vector $V'_j$, and $\hat{PBI}_j(\bm{x})$ is the candidate's projected PBI value under the same vector. In the design of AdaE-SAEA, we adopt two variants of $\text{ND-}\Delta PBI$ by adjusting the penalty coefficient $\lambda$, which controls the trade-off between convergence and diversity. Specifically, we set $\lambda = 3$ to emphasize convergence by prioritizing the projection distance along the reference direction, and $\lambda = 8$ to emphasize diversity by amplifying the perpendicular deviation.

EPDI is a model uncertainty-aware infill criterion designed to balance exploration and exploitation in multi-objective optimization. It extends the classical EI concept to a vector-valued setting by introducing the Proximity and Diversity scalarization function, which simultaneously captures convergence and diversity characteristics. Let $\hat{F}(\bm{x})$ denote the predicted objective vector of solution $\bm{x}$, and let $V_r$ be a randomly sampled reference vector. The PD function is defined as:
\begin{equation}
PD(\hat{F}(\bm{x}), V_r) = \frac{1}{m} \sum_{i=1}^m \hat{f}_i(\bm{x}) + \lambda \left| \hat{F}(\bm{x}) \right|_2 \cdot \sin(\hat{F}(\bm{x}), V_r),
\end{equation}
where the first term represents the average convergence of $\bm{x}$ across $m$ objectives, and the second term penalizes angular deviation from the reference direction $V_r$ to promote diversity. The sine term is used to measure the vertical (perpendicular) distance between $\hat{F}(\bm{x})$ and $V_r$ in the objective space. Given a solution archive $A_{\text{rnd}}$, the improvement function PDI is computed as:
\begin{equation}
PDI(\hat{F}(\bm{x}), V_r) = \max\left(PD_{\min} - PD(\hat{F}(\bm{x}), V_r),\ 0\right),
\end{equation}
where $PD_{\min}$ is the minimal PD value among all solutions in the archive associated with $V_r$. Finally, EPDI computes the expectation of improvement under model uncertainty via:
\begin{align}
\begin{aligned}
EPDI(\bm{x}, &\hat{F}, V_r) =
&\int_{y \in \mathbb{R}^m} PDI(\hat{F}(\bm{x}), V_r) \prod_{i=1}^m \frac{1}{\delta_i} \phi\left(\frac{y_i - \hat{y}_i}{\delta_i}\right) dy_i,
\end{aligned}
\end{align}
where $\phi(\cdot)$ is the standard normal density, $\hat{y}_i$ and $\delta_i$ denote the predicted value and standard deviation for objective $i$, respectively. In practice, the integral is approximated via Monte Carlo sampling with 1000 draws. The candidate with the highest EPDI value is selected. In AdaE-SAEA, we design two EPDI variants by adjusting the trade-off coefficient $\lambda = 3$. When $\lambda = 3$, the selection function emphasizes exploitation by prioritizing solutions with better predicted convergence. In contrast, setting $\lambda = 8$ amplifies the importance of diversity, encouraging exploration toward less-visited directions in the objective space. This adaptive configuration enables AdaE-SAEA to balance global exploration and local exploitation effectively across different optimization stages.

\section{Exploratory Landscape Analysis Features}
\label{app2}
As mentioned in Section \ref{subsubsec4.3.1}, ELA is widely used for instance space analysis in BBO problems. In this study, we selected 15 ELA features in total (see Table \ref{tab:b1}). Among them, $s_1 \text{-} s_{10}$ were primarily inspired by the R2-RLMOEA framework \cite{a44}. Specifically, $s_1 \text{-} s_6$ describe the population performance statistics of the current generation. By scalarizing individual performance through the R2 indicator and computing quartiles relative to minimum and maximum values, these features reflect overall convergence tendency and population dispersion. $s_7 \text{-} s_{10}$ capture geometric structure in the decision space, computed as the Euclidean distances between $Q_1/Q_2/Q_3/Q_{mean}$ quartile points and the best individual, normalized by the maximal decision span. This allows the ELA to perceive the spatial distribution and morphological evolution of the search process.

Then, We further introduce three state features from DRL-SAEA \cite{29} as $s_{11}$–$s_{13}$: convergence  $(s_{11})$, diversity $(s_{12})$, and evolutionary stage $(s_{13})$, to enhance the ELA’s ability to perceive global search information. Convergence measures the proximity of the population’s average objective values to the Pareto optimal front, reflecting the overall optimization progress. Diversity characterizes the spread of solutions in the objective space, where higher diversity indicates broader exploration and helps prevent premature convergence. The evolutionary stage represents the proportion of evaluations used relative to the total evaluation budget, capturing the temporal position of the search process. These three features describe the optimization status from complementary perspectives of global convergence, solution distribution, and search progression, enabling the ELA to distinguish between exploration and exploitation phases and providing the reinforcement learning agent with a richer and more temporally aware state representation.

\begin{table}[H]  
  \centering
  \caption{State definition and explanation}
  \label{tab:b1}
  \begin{adjustbox}{max width=\textwidth}
  \renewcommand{\arraystretch}{1.3}
  \small  
  \begin{tabular}{clp{8cm}} 
    \toprule
    \textbf{Index} & \textbf{State} & \textbf{Explanation} \\
    \midrule
    $S_1$ & $\dfrac{f_{Q1} - f_{\min}}{f_{\max} - f_{\min}}$ &
    \makecell[lp{8cm}]{
    $f_{Q1}$: lower quartile of population performance \\
    $f_{\min}$: minimum performance \\
    $f_{\max}$: maximum performance} \\

    $S_2$ & $\dfrac{f_{Q2} - f_{\min}}{f_{\max} - f_{\min}}$ & 
    \makecell[lp{8cm}]{$f_{Q2}$: median of population performance.} \\

    $S_3$ & $\dfrac{f_{Q3} - f_{\min}}{f_{\max} - f_{\min}}$ & 
    \makecell[lp{8cm}]{$f_{Q3}$: upper quartile of population performance} \\

    $S_4$ & $\dfrac{f_{Qmean} - f_{\min}}{f_{\max} - f_{\min}}$ & 
    \makecell[l]{$f_{Qmean}$: average of all quartiles} \\

    $S_5$ & $\dfrac{SD(\mathcal{P})}{SD(\mathcal{P}_{\min \max})}$ & 
    \makecell[lp{8cm}]{
        $SD(\mathcal{P})$: population performance standard deviation \\
        $SD(\mathcal{P}_{\min \max})$: maximum performance standard deviation 
        (half of $\mathcal{P}_{\min \max}$ includes minimum performance and the rest maximum performance)
        } \\

    $S_6$ & $\dfrac{G_{\max} - G_t}{G_{\max}}$ & 
    \makecell[lp{8cm}]{
    $G_{\max}$: maximum generation \\
    $G_t$: current generation} \\

    $S_7$ & $\dfrac{ED(x_{Q1}, x_{\min})}{ED(x_{\max}, x_{\min})}$ & 
    \makecell[lp{8cm}]{
    $ED(\cdot)$: Euclidean distance} \\
    
    $S_8$ & $\dfrac{ED(x_{Q2}, x_{\min})}{ED(x_{\max}, x_{\min})}$ & 
    \makecell[lp{8cm}]{
    $x_{\min}$: decision variables related to $f_{\min}$ \\
    $x_{\max}$: decision variables related to $f_{\max}$ \\} \\

    $S_9$ & $\dfrac{ED(x_{Q3}, x_{\min})}{ED(x_{\max}, x_{\min})}$ & 
    {} \\

    $S_{10}$ & $\dfrac{ED(x_{Qmean}, x_{\min})}{ED(x_{\max}, x_{\min})}$ & 
    {} \\

    $S_{11}$ & $\sum_{i=1}^m\sum_{\bm{x}\in P}\frac{f_i{(\bm{x})}}{N}$ & 
    \makecell[lp{8cm}]{
    $P$: current population \\
    $N$: population size \\
    } \\

    $S_{12}$ & $\sum_{i=1}^m\sum_{\bm{x}\in P}\frac{{(f_i{(\bm{x})}}-obj_i)^2}{N}$ & 
    \makecell[lp{8cm}]{
    $obj_i:\text{}\sum_{\bm{x}\in P} f_i(\bm{x})/N$
    } \\

    $S_{13}$ & $\frac{FE_t}{FE_{max}}$ & 
    \makecell[lp{8cm}]{
    $FE_t$: number of evaluations used \\
    $FE_{max}$: maximum number of evaluations\\
    } \\

    $S_{14}$ & $\frac{1}{N}\sum_{i=1}^m\sum_{x \in P}|\frac{f_i(x)-\hat{f_i}_{\text{Bagging}}(x)}{f_i(x)+\epsilon}|$ & 
    \makecell[lp{8cm}]{
    $\epsilon:0.01$ \\
    $\hat{f_i}_\text{Bagging}(\bm{x})$: Bagging surrogate prediction value \\
    } \\

    $S_{15}$ & $\frac{1}{N}\sum_{i=1}^m\sum_{x \in P}|\frac{f_i(x)-\hat{f_i}_{\text{Boosting}}(x)}{f_i(x)+\epsilon}|$ & 
    \makecell[lp{8cm}]{
    $\hat{f_i}_\text{Boosting}(\bm{x})$: Boosting surrogate prediction value
    } \\

    \bottomrule
  \end{tabular}
  \end{adjustbox}
\end{table}

Finally, we incorporate the predictive accuracy of surrogate models into the ELA feature set by introducing $s_{14}$ and $s_{15}$, which correspond to the MAPE of Bagging and Boosting surrogate modeling, respectively. These features measure how well the surrogate models fit the already evaluated population, providing a dynamic indicator of the approximation quality between the surrogate and the true objective functions. By integrating $s_{14} \text{-} s_{15}$, the ELA not only captures convergence and distribution information but also directly reflects surrogate model performance, enabling the reinforcement learning agent to select appropriate modeling strategies at different stages of the optimization process.

\bibliographystyle{unsrt}
\bibliography{reference}

\end{document}